# LAD-RCNN:A Powerful Tool for Livestock Face Detection and Normalization

Ling Sun [a, b], Guiqiong Liu [a, b], Xunping Jiang [a, b, c, *], Junrui Liu [b], Xu Wang [b], Han Yang [b], Shiping Yang [b]

## Abstract

With the demand for standardized large-scale livestock farming and the development of artificial intelligence technology, a lot of research in area of animal face recognition were carried on pigs, cattle, sheep and other livestock. Face recognition consists of three sub-task: face detection, face normalization and face identification. Most of animal face recognition study focuses on face detection and face identification. Animals are often uncooperative when taking photos, so the collected animal face images are often in arbitrary directions. The use of non-standard images may significantly reduce the performance of face recognition system. However, there is no study on the normalization of the animal face image with arbitrary directions. In this study, we developed a light-weight angle detection and region-based convolutional network (LAD-RCNN) containing a new rotation angle coding method that can detect the rotation angle and the location of animal face in one-stage. LAD-RCNN has been evaluated on multiple dataset including goat dataset and gaot infrared image, and has a frame rate of 72.74 FPS (including all steps) on a single GeForce RTX 2080 Ti GPU. Evaluation result show that the AP of face detection was more than 95% and the deviation between the detected rotation angle and the ground-truth rotation angle were less than 6.42° on all the test dataset. This shows that LAD-RCNN has excellent performance on livestock face and its direction detection, and therefore it is very suitable for livestock face detection and Normalizing. Code is available at https://github.com/SheepBreedingLab-HZAU/LAD-RCNN/

## Keywords

Livestock face detection; Rotation angle detection; Livestock face normalization; Face recognition ;

[*] Corresponding author: Dr. Xunping Jiang, xpjiang@mail.hzau.edu.cn
[a] Key Laboratory of Smart Farming for Agricultural Animals, Wuhan 430070, People's Republic of China
[b] Laboratory of Small Ruminant Genetics, Breeding and Reproduction, College of Animal Science and Technology, Huazhong Agricultural University, Wuhan 430070, People's Republic of China.
[c] Key Laboratory of Agricultural Animal Genetics, Breeding and Reproduction of the Ministry of Education, Wuhan 430070, People's Republic of China.

# 1  Introduction

The need for identification of individual livestock has become an urgent problem with the requirement of quality control, welfare management, and large-scale livestock farming of livestock[1]. Ear tags and Radio Frequency Identification (RFID) tags are currently commonly used for livestock individual identification, but those tags need to be nailed to the ears or implanted under the skin of livestocks, which may reduce welfare of livestocks, and RFID tags can only be readed when they are close to RFID reader [2]. In contrast, face recognition technolgoy can realize non-contact livestock identification, which can reduce animal stress. Animal face recognition technology has been widely studied in animal husbandry [3], especially in the field of pig, cattle and sheep, and goat face recognition [1, 4, 5].

Face recognition consists of three sub-tasks: face detection, face normalizing, and face identification [6, 7]. Among them, face detection is to detect the location of face in image; Face normalizing is to align the faces to normalized coordinates; Face identification is implemented on normalized face[6]. Most of livestock face recognition study focuses on face identification, and there are some studies focus on face detection, such as Billah et al [4] detected the goat face location with YOLO V4; Hitelman et al [8] detect the sheep face location through Faster RCNN; Wang and Liu [9] detected pig face location through EfficientDet-D0. However, no study was found on livestock face normalizing.

Face recognition without face normalization will significantly reduce the performance [10]. The way of human face normalization is mainly through keypoint detection [7], and the face image is normalized by affine transformations such as rotation and scaling according to the location of detected keypoints (Zhang et al., 2016). The keypoint detection technology has well performance in human face normalization. However, faces of arbitrary direction may be captured and need to be recognized in field of livestock face recognition (Fig. 1), and the keypoint detection may failed when the face rotation angle is too large tested on dlib's shape detector. In order to realize the normalization of livestock face in arbitrary direction, we propose a Light-weight Angle Detection and Region-based Convolutional Network (LAD-RCNN). LAD-RCNN realizes face location detection and face rotation angle detection in one-stage. Livestock face normalization will be conveniently achieved through clipping, rotation and scaling according to the face location and rotation angle detected by LAD-RCNN.

We summarize our contributions as follows:

（1） We propose LAD-RCNN for livestock face detection and normalization, which can handle arbitrary direction of livestock face.
（2） We propose a light-weight backbone for LAD-RCNN, which is faster than MobileNetV2, ResNet50, and VGG16 with no significant accuracy reduction on LAD-RCNN.
（3） The average detection speed of LAD-RCNN reaches 72.74 FPS tested on a single GeForce RTX 2080 Ti GPU.
（4） LAD-RCNN was evaluated in multiple dataset and the AP was more than

95%, the average angle difference between the detection angle and the ground-truth angle was within 6.42°.

## 2 Related Work

### 2.1 Object Detection

Object detection can be classified into two categories: "one-stage detection" and "two-stage detection". The one-stage detection has no region proposal stage and detect the location and classification in one-stage; the two-stage detection carries out region proposal first, and then carries out classification and location detection [11]. Faster R-CNN [12] and Mask RCNN [13] are currently widely used two-stage object detection methods. Faster R-CNN is developed on the basis of Fast R-CNN [14] by replacing time-consuming selective search with region proposal network (RPN) to improve detection speed. Mask R-CNN [13] adds a branch for predicting an object mask in parallel with the existing bounding box recognition branch in Faster R-CNN to realize the segmentation task, and replaces ROI Pooling by ROI Align to improve the performance of the segmentation task. SSD[15] is the first widely used one-stage object detection method, which uses multi-scale feature maps to detect object of different size. RetinaNet [16] introduced focal loss to solve the problem of "imbalance between positive and negative samples" in one-stage object detector, which improves the detection accuracy; YOLO V4 [17] is a commonly used one-stage detection model. YOLO V4 is developed based on previous version of YOLO models [18-20] and introduces a series of features to increase detection accuracy. In recent years, it has also been reported that the transformers [21], which has been widely used in Natural Language Processing (NLP), has been used for object detection [22, 23].

### 2.2 Angle-based Rotated Object Detection

Angle-based rotated object detection method have developed rapidly in aerial object detection and text detection, and it is developed by add an angle detection into the object detection, and usually represented as vector ($x$, $y$, $w$, $h$, $\theta$) [24]. Since the performance of the two-stage detector performance better than one-stage detector in rotated object detection, most of the rotated object detector relies on the two-stage RCNN frameworks by replacing anchors and RoI pooling with Rotation anchors and rotated RoI pooling [25-27]. The character of aerial images are small and densely packed, which is hard to detect [28]. In order to have better performance in aerial object detection, R$^2$PN [29] generates anchors in multiple directions by controlling of scale, rations and angle, and redefines the IoU computation method. SCRNet [30] proposes a multi-dimensional attention network to reduce noise interference and improve the sensitivity to small objects, and add IoU constant factor to loss function, so that the loss function can better handle rotating bounding box regression. Yang and Yan [31] deal with angle prediction question as classification to alleviate the discontinuous boundary problem and proposed circular smooth label (CSL) technology to detect large aspect ratio object. ReDet [32] proposes a rotation-equivariant backbone to extract rotation-equivariant features and a rotation-invariant RoI Align to get rotation-invariant features, which reduce the number of parameters. Oriented R-CNN [33] proposes a oriented Region Proposal Network (oriented RPN) that generates oriented proposals on less anchor, which improves the detection speed. MRDet [27] proposes an arbitrary-oriented Region Proposal Network (AO-RPN) that add a branch in PRN to learn transformations parameters for generating oriented proposals; SCRDet++[34] extends SCRDet through instance level denoising modules to improve the performance of small and densely object detection.

These angle-based rotated object detector have excellent performance in aerial object detection and text detection. However, the rotation angle is represented as the angle between the long axis and the horizontal axis in those study[24], which may get reversed result in normalizing livestock face (Supplementary Fig. 1). Therefore, it is necessary to design new rotation angle representation methods, and then propose a new angle detection and region-based convolutional network suitable for livestock face normalization.

# 3 APPROACH

## 3.1 Model

### 3.1.1 Rotation angle

We use the angle between the horizontal axis and the line from left key-point to right key-point to represent the rotation angle of object. These two key-points can be selected empirically by the principle that the line from left key-point to right should be parallel to the horizontal axis in the standardized object. We use floating number between (-1.0, 1.0] to represent rotation angle between (-180°, 180°], where a positive value indicates counterclockwise rotation, and a negative value indicates clockwise rotation. The calculation method of rotation angle ($\theta$) is shown in formula 1, formula 2 and figure 1.

$$k = \begin{cases} \dfrac{y_l - y_r}{x_r - x_l} & (x_r \neq x_l) \\ (y_{l-}y_r) \times \infty & (x_r = x_l) \end{cases} \quad (1)$$

$$\theta = \begin{cases} \dfrac{\tan^{-1}(k)}{\pi} & (x_r - x_l > 0) \\ 0.5 & (y_l - y_r > 0, x_r - x_l = 0) \\ 1 - \dfrac{\tan^{-1}(|k|)}{\pi} & (y_l - y_r \geq 0, x_r - x_l < 0) \\ -0.5 & (y_l - y_r < 0, x_r - x_l = 0) \\ \dfrac{\tan^{-1}(|k|)}{\pi} - 1 & (y_l - y_r < 0, x_r - x_l < 0) \end{cases} \quad (2)$$

Where, $x_l$ is the distance between the left key-point and the left frame of the picture; $x_r$ is the distance between the right key-point and the left frame of the picture; $y_l$ is the distance between the left key point and the upper frame of the picture; $y_r$ is the distance between the right key point and the top frame of the picture.

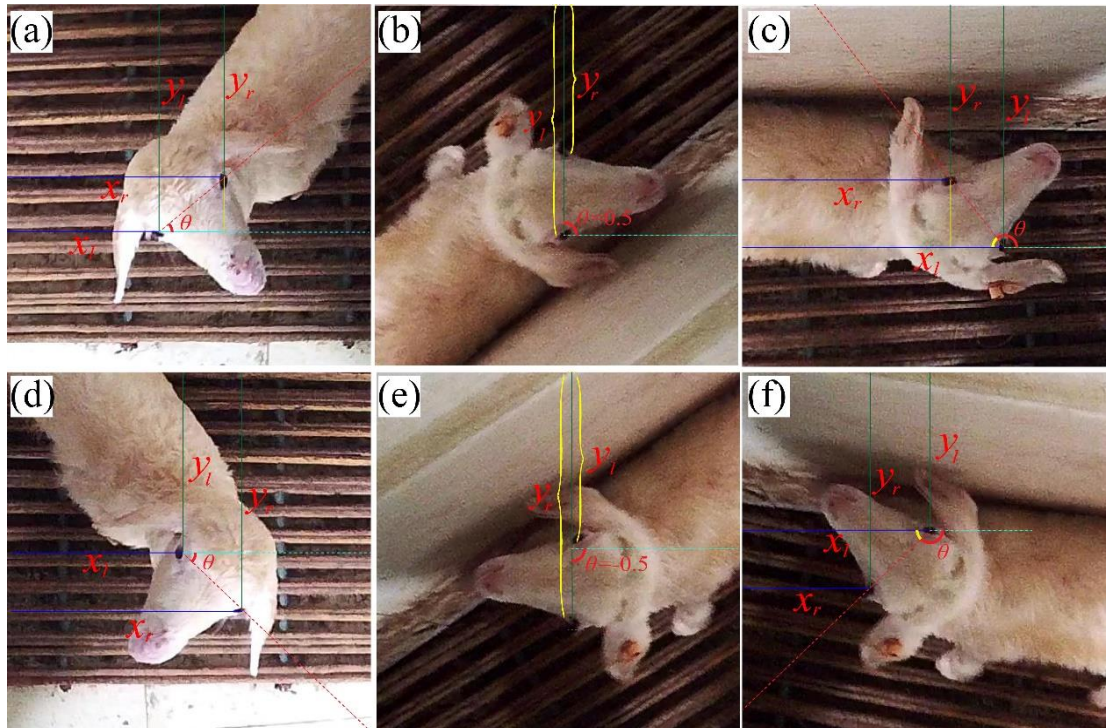

Fig. 1. Illustration of angle definition. We choose the left eye in the normalized picture as the left key-point, and the right eye as the right key-point; (a) and (d) correspond to the condition $(x_r - x_l > 0)$, where $\theta$ is between [0, 0.5) in (a) and between (0, -0.5) in (d); (b) correspond to the condition $(y_l - y_r > 0, x_r - x_l = 0)$, where $\theta = 0.5$; (c) correspond to the condition $(y_l - y_r \geq 0, x_r - x_l < 0)$, where $\theta$ is between (0.5, 1]; (e) correspond to the condition $(y_l - y_r < 0, x_r - x_l = 0)$, where $\theta = -0.5$; (f) correspond to the condition $(y_l - y_r < 0, x_r - x_l < 0)$, where $\theta$ is between (-0.5, -1).

### 3.1.2    Angle discontinuity problem

The difference of rotation angle is little between the object rotating counterclockwise by nearly 180° ($\theta \rightarrow 1.0$) and the object rotating clockwise by nearly 180° ($\theta \rightarrow -1.0$), but the difference of the calculated $\theta$ is very large (Fig. 2). It may cause the model not to converge in training. To deal with this problem, we split the angle value $\theta$ into its absolute value and its sign on the reson of that its absolute value are continuous. Therefore, LAD-RCNN detect the absolute value and the direction of rotation angle, respectively.

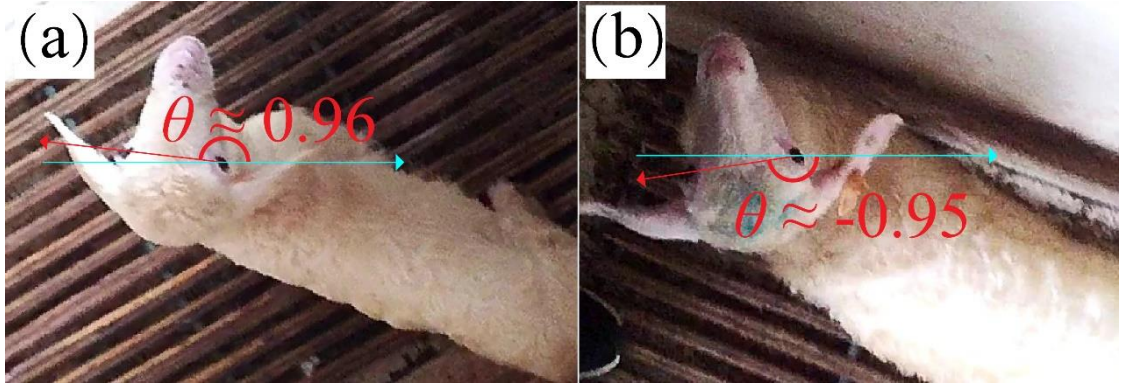

Fig. 2. Angle discontinuity problem. The difference of rotation angles in (a) and (b) is little, but the difference of calculated $\theta$ is very large.

### 3.1.3 Overall structure

The overall structure of LAD-RCNN is depicted in Fig. 3. Three tensors are generated from input image by backbone network. After convolution, up sampling, and addition operations, four feature maps with different sizes are generated from those three tensors. The four feature maps are convoluted with the same kernel to output tensors with 54 channels (each cell corresponds to 6 anchors, and each anchor corresponds to 9 numbers). Output tensors generated from four feature maps are concatenated and reshape to a tensor with 9 channels. Among them, 2 channels are used for objectness detection, 4 channels are used for box encodings detection, 2 channels are used for angle direction detection, and 1 channel is used for angle value detection.

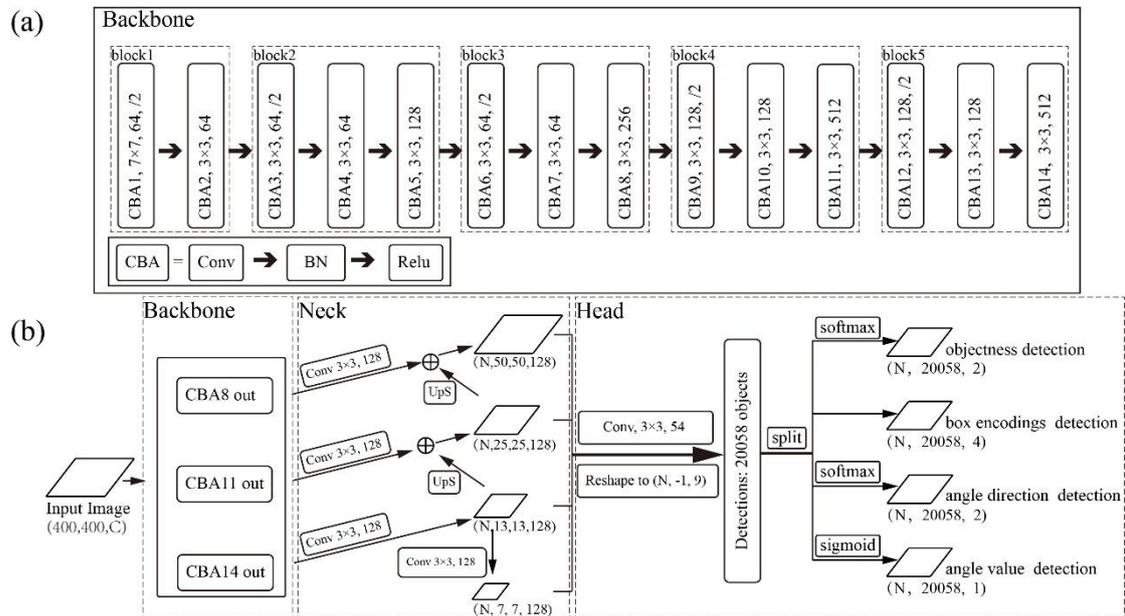

Fig. 3. The overall pipeline of the LAD-RCNN. (a): backbone network; (b): overall pipeline of the LAD-RCNN; The rounded rectangle represents the operation on the tensor; The rhombus represents the tensor; CBA represents the sequential operation of convolution, normalization and relu activation; $\oplus$ represents add operation; UpS indicates the up-sampling operation.

### 3.1.4 Backbone

The backbone network consists of 14 sequential CBA, and each CBA consists of a convolution layer, a batch normalization layer, and an activation layer (Fig. 3.). The first CBA uses 7×7 kernels in convolution, and the other CBA use 3×3 kernels. The backbone network is divided into 5 blocks. In each block, the step length is 2 in first convolution layer. In the last four blocks, the dimension of output tensor of the third CBA is four times the size of first two CBA. Output tensor of last CBA in last three blocks is transferred to the neck network.

### 3.1.5 Anchor

We set a series of anchors associated with cell in feature maps inspired by [15]. The center point of the anchor is determined by the position of associated cell. The initial size of the anchor is preset, and the anchor associated with the shallow feature is smaller than the deep feature. Each cell in feature maps is associated with $k$ ($k$=6 in default) anchors. We set 6 anchors per cell controlled by 2 scaling ratios $(1, \sqrt{2})$ and 3 aspect ratios $(1.0, 2.0, 0.5)$ in default.

### 3.1.6 Head Network

Each anchor box corresponds to nine values (Fig. 3.), in which four values are used to detect box, two values are used to detect objectness, two values are used to detect rotation direction, and one value is used to detect absolute angle value.

We decode the box prediction result through bounding box regression [12, 35]:

$$
\begin{aligned}
x &= (t_x/10.0) \times w_a + x_a & y &= (t_y/10.0) \times h_a + y_a \\
w &= e^{(t_w/5.0)} \times w_a & h &= e^{(t_h/5.0)} \times h_a
\end{aligned}
\tag{3}
$$

Where, $x$, $y$, $w$, and $h$ denote the predicted box's center coordinates and its width and height. $t_x$, $t_y$, $t_w$, and $t_h$ denote the output tensor of CNN. $x_a$, $y_a$, $w_a$, and $h_a$ denote the anchor's center coordinates and its width and height.

Accordingly, we encode the ground-truth box as follows:

$$
\begin{aligned}
t_x^* &= 10.0 \times \frac{(x^* - x_a)}{w_a}, & t_y^* &= 10.0 \times \frac{(y^* - y_a)}{h_a}, \\
t_w^* &= 5.0 \times log\left(\frac{w^*}{w_a}\right), & t_h^* &= 5.0 \times log\left(\frac{h^*}{h_a}\right)
\end{aligned}
\tag{4}
$$

Where, $x^*$, $y^*$, $w^*$, and $h^*$ denote the ground-truth box's center coordinates and its width and height.

We convert the objectness detection and rotation direction detection result through SoftMax function.

$$
Softmax(z_j) = \frac{e^{z_j}}{\sum_i e^{z_i}}
\tag{5}
$$

Where, $i \: and \: j \in \{0, 1\}$, $z_j$ denote the $j$-th value. $Softmax(z_j)$ denote the calculated probability.

We convert the angle value result through sigmoid function:

$$
Sigmoid(x) = \frac{1}{1 + e^{-x}}
\tag{6}
$$

## 3.2 Training

### 3.2.1 Dual dataset training

LAD-RCNN is trained by datasets both with angle data (dataset 1) and without angle data (dataset 2). Dataset 1 is mainly used to train the rotation angle and rotation direction; Dataset 2 is mainly used for objectness detection and box encodings detection.

### 3.2.2 Loss Function

The overall loss function of LAD-RCNN are weighted sum of object localization loss, objectness loss, absolute angle value loss and angle direction loss:

$$L_{LAD-RCNN} = \lambda_{loc}L_{loc} + \lambda_{obj}L_{obj} + \lambda_{av}L_{av} + \lambda_{ad}L_{ad} \tag{7}$$

Where $\lambda_{loc}$, $\lambda_{obj}$, $\lambda_{av}$ and $\lambda_{ad}$ are the trade-off parameters and are set to 1.0, 5.0, 1.0 and 10.0 by default, respectively. $L_{loc}$ denotes localization loss; $L_{obj}$ denotes objectness loss; $L_{av}$ denotes absolute angle value loss; $L_{ad}$ denotes angle direction loss.

We employ mini-batch sampling [14] to deal with imbalance between positive and negative samples in training. Localization loss is defined as:

$$L_{loc} = \frac{1}{N_{loc}} \sum_{i=1}^{N_{loc}} \sum_{j \in \{x,y,w,h\}} \text{Huber}(t^*_{i,j} - t_{i,j}) \tag{8}$$

In which,

$$\text{Huber}(a) = \begin{cases} 0.5a^2 & (|a| \leq \delta) \\ \delta|a| - 0.5\delta^2 & (|a| > \delta) \end{cases} \tag{9}$$

Here, $N_{loc}$ is the number of positive anchor in mini-batch, and $i$ is the index of positive anchor in mini-batch; $x$, $y$, $w$, and $h$ are the same with formula 1. $t^*_{i,j}$ is ground-truth value of $j$ corresponding to the $i$-th anchor calculated by formula 1; $t_{i,j}$ is the predicted value of $j$ corresponding to the $i$-th anchor calculated by formula 1; $\delta$ is a variable in huber function and we set $\delta=1$ in default.

Objectness loss is defined as:

$$L_{obj} = \frac{1}{N_{obj}} \sum_{i=1}^{N_{obj}} \text{FL}(1 - |p_i - p^*_i|) \tag{10}$$

In which,

$$\text{FL}(p_t) = -(1 - p_t)^\gamma \log(p_t) \tag{11}$$

Here, $N_{obj}$ is the number of anchors in mini-batch, the $i$ is the index of anchor in mini-batch, $p_i$ is the predicted probability that the $i$-th anchor is marked as an object; $p^*_i$ indicating whether the $i$-th anchor box is marked as an object. When the $i$-th anchor is marked as an object, $p^*_i = 1$, otherwise, $p^*_i = 0$. FL(*) is focal loss function[16], and we set $\gamma=2$ by default.

Absolute angle value loss is defined as:

$$L_{av} = \frac{1}{N_{loc}} \left( \sum_{i=1}^{N_{loc}^{ds1}} \lambda_{ds1} \times 0.5(\theta^*_{v,i} - \theta_{v,i})^2 + \sum_{i=1}^{N_{loc}^{ds2}} \lambda_{ds2} \times 0.5(\theta_{v,i})^2 \right) \tag{12}$$

Here, $N_{loc}$ is the number of positive anchor in mini-batch, and $i$ is the index of positive anchor in mini-batch; $N_{loc}^{ds1}$ is the number of positive anchor corresponding to dataset 1 which with angle data. $N_{loc}^{ds2}$ is the number of positive anchor corresponding to dataset 2 which without angle data. $\theta_{v,i}$ is the predicted absolute angle value of the $i$-th anchor; $\theta_{v,i}^*$ is the ground-truth absolute angle value of the $i$-th anchor; $\lambda_{ds1}$ and $\lambda_{ds2}$ are the trade-off parameters and are set to 10.0 and 0.0 by default, respectively.

Angle direction loss is defined as:

$$L_{ad} = \frac{1}{N_{ad}} \sum_{i \in I} FL(1 - |p_{\theta,i} - p_{\theta,i}^*|) \tag{13}$$

In which,

$$I = \{i | i \in A, |\theta_{v,i}^*| > \varepsilon\} \tag{14}$$

Here, $N_{ad}$ is the number of elements in set $I$. FL (*) is the focal loss function defined by formula 5; $p_{\theta,i}$ is the predicted probability that the $i$-th anchor is counterclockwise rotation; $p_{\theta,i}^*$ indicating the ground-truth probability of whether the $i$-th anchor box is counterclockwise rotation. When the $i$-th anchor marked as counterclockwise rotation, $p_{\theta,i}^* = 1$, otherwise, $p_{\theta,i}^* = 0$; $A$ is the set of all anchors; $\theta_{v,i}^*$ is the ground-truth absolute angle value of the $i$-th anchor; $\varepsilon$ is a preset parameter with default value of 0.025.

### 3.2.3 Data Augmentation

To make LAD-RCNN more robust to arbitrary rotation angle and suitable for small dataset, the training set can be randomly operated by following operations:

**Counterclockwise rotation by 90°.** The ground-truth angle after rotation can be calculated as follows:

$$\theta' = \begin{cases} \dfrac{\theta \times 180 + 90}{180} & (\theta \leq 0.5) \\ \dfrac{\theta \times 180 - 270}{180} & (\theta > 0.5) \end{cases} \tag{15}$$

Where, $\theta$ is the original angle, $\theta'$ is the angle after operation, the same below.

**Horizontally flipping.** The ground-truth angle after horizontally flipping can be calculated as follows:

$$\theta' = -\theta \tag{16}$$

**Vertically flipping.** The ground-truth angle after vertically flipping can be calculated as follows:

$$\theta' = \begin{cases} |\theta| - 1 & (\theta < 0) \\ 1 - \theta & (\theta \geq 0) \end{cases} \tag{17}$$

**Image tiling.** During the training, the images are tiled together with a preset probability. The images is tiled by combining 4 images into 1 image in the form of 2×2, or combining 9 images into 1 image in the form of 3×3.

These operations are independent of each other, and trigger probability can be set separately for each operation. Thus, a image may be operated in multiple ways in a training.

# 4 Evaluation

## 4.1 Evaluation metrics

The performance of LAD-RCNN was measured by precision, recall rate, F1-score, average precision, and average angle difference (AAD).

$$Precision = \frac{TP}{TP + FP} \quad (18)$$

$$Recall = \frac{TP}{TP + FN} \quad (19)$$

$$F_1 - score = 2 \times \frac{Precision \times Recall}{Precision + Recall} \quad (20)$$

Where *TP*, *FP*, and *FN* are the number of true positive, false positive, and false negative prediction box at IoU=0.5, respectively.

AP is the area under the precision-recall curve, which is widely used for in object detection evaluation [36], the calculating formula is as follows:

$$AP = \int_0^1 p(r)\, dr \quad (21)$$

Where, *r* represents recall rate, *p(r)* is the precision when the recall rate is *r*.

The performance of angle detection of LAD-RCNN was measured by average angle difference(*AAD*) between the detection angle and the ground-truth angle:

$$AAD = \frac{1}{N_{obj}} \left( \sum_{i=1}^{N_{obj}} D(\theta_i^*, \theta_i) \right) \times 180° \quad (22)$$

In which,

$$D(\theta_i^*, \theta_i) = \begin{cases} |\theta_i^* - \theta_i| & (|\theta_i^* - \theta_i| < 1) \\ 2 - |\theta_i^* - \theta_i| & (|\theta_i^* - \theta_i| \geq 1) \end{cases} \quad (23)$$

Where, $N_{obj}$ is the total number of objects detected in test set. $\theta_i^*$ is the ground-truth angle with direction corresponding to *i*-th detected objects; $\theta_i$ is the predicted angle with direction in the *i*-th detected objects.

## 4.2 backbone evaluation

LAD-RCNN consists of backbone, neck network and head network (Fig. 3). In addition to our backbone, LAD-RCNN also supports the use of other backbone network, such as MobileNetV2, VGG16 and ResNet50, etc. MobileNetV2 [37] is a lightweight network designed for mobile user; VGG16 is a classic backbone network; ResNet50 is a widely used deep convolutional network.

Table 1. Comparison of backbones

| Backbone | Input resolution | Parameters | FPS |
|---|---|---|---|
| Ours | 400×400 | 2.82M | 72.74 |
| MobileNetV2 | 400×400 | 2.26M | 53.37 |
| VGG16 | 400×400 | 14.71M | 55.04 |
| ResNet50 | 400×400 | 23.59M | 44.32 |

Note: FPS is the test result including all steps of LAD-RCNN on a single RTX 2080Ti GPU.

Table 1 shows the comparison between our backbone, MobileNetV2, VGG16 and ResNet50. The results show that the number of parameters in our backbone is far less than that of VGG16 and ResNet50, which is like the lightweight network MobileNetV2, and the detection speed of LAD-RCNN with our backbone is 36.29%, 32.15% and 64.12% faster than that of LAD-RCNN with MobileNetV2, VGG16 and ResNet50, respectively.

## 4.3 Experiments on goat dataset

The goat dataset [4] labeled the location of the goat face and eyes, which contains 1680 training data and 1311 test data. There are 438 and 613 images containing two eyes in training set and test set, respectively. The face rotation angle was calculated according to the location of two eyes. The dataset 1 containing angle information was generated by training data containing rotation angle. The dataset 2 without angle information was generated by all the training data.

The training parameters were set as follows: the probabilities of horizontal flip, vertical flip and 90° rotation of dataset 1 were set to 0.5; The probability of horizontal flip and vertical flip of dataset 2 were set to 0.5, and probability of 90° rotation of dataset 2 does was set to 0; The probability of 2×2 merger in dataset 1 and dataset 2 were set to 0.8; The batchsize of dataset 1 was set to 7, and that of dataset 2 was set to 5; The input image channel was set to 3; The total training step was set to 50000.

Table 2. Test result of LAD-RCNN on goat dataset

| Backbone | Precision | Recall | F1 score | AP | AAD |
|---|---|---|---|---|---|
| Ours | 95.02% | 90.70% | 92.81% | 97.55% | 6.42° |
| MobileNetV2 | 89.23% | 90.30% | 89.76% | 95.25% | 4.98° |
| VGG16 | 64.89% | 79.67% | 71.52% | 79.80% | 9.08° |
| ResNet50 | 88.99% | 91.64% | 90.30% | 95.62% | 6.12° |

Note: AAD represents average angle difference between the detection angle and the ground-truth angle.

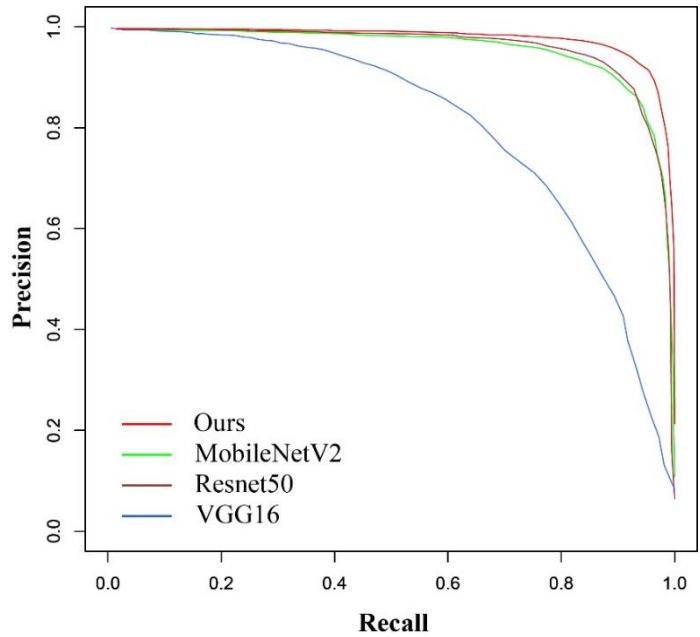

Fig. 4. Precision-Recall Curves on goat dataset.

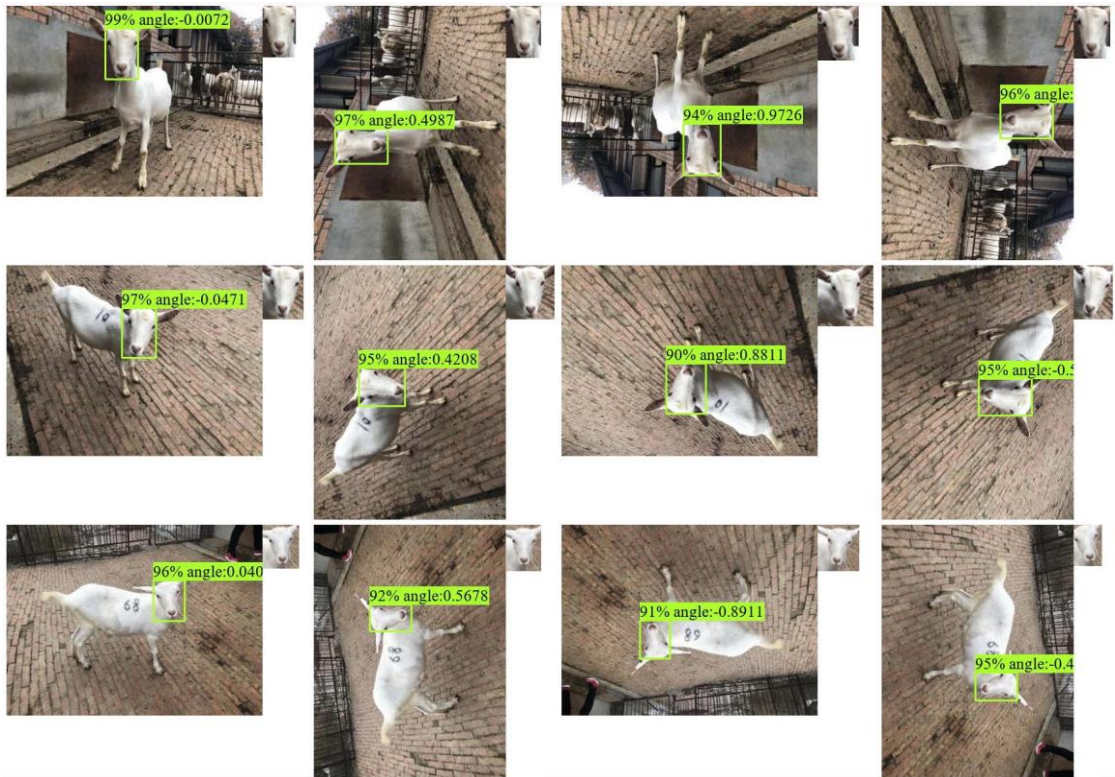

Fig. 5. Detection examples on goat image with LAD-RCNN. The small image in the upper right corner of each image is the extracted normalized face according to the detection result. The four pictures in each line represent the same picture in test set, which are original image and images rotated by 90°, 180 ° and 270°, respectively

The 613 images in test data containing angle information were used to evaluate the trained model. To evaluate the performance of LAD-RCNN on detecting goat face with arbitrary rotation angles, we rotated the test image by 90°, 180° and 270° respectively to form a new test dataset with 613 × 4 images. The test results (Table 2, Fig. 4, and Fig. 5) show that the AP were more than 95% when Ours, MobileNetV2 or

ResNet50 were adopted as the backbone network, and the AP was the highest when our backbone was adopted. When Ours, MobileNetV2 or ResNet50 were used as the backbone network, the average angle difference were within 6.42°.

The model trained by the goat dataset also performs well in detecting and normalizing face in sheep bird-view image (Fig. 6)

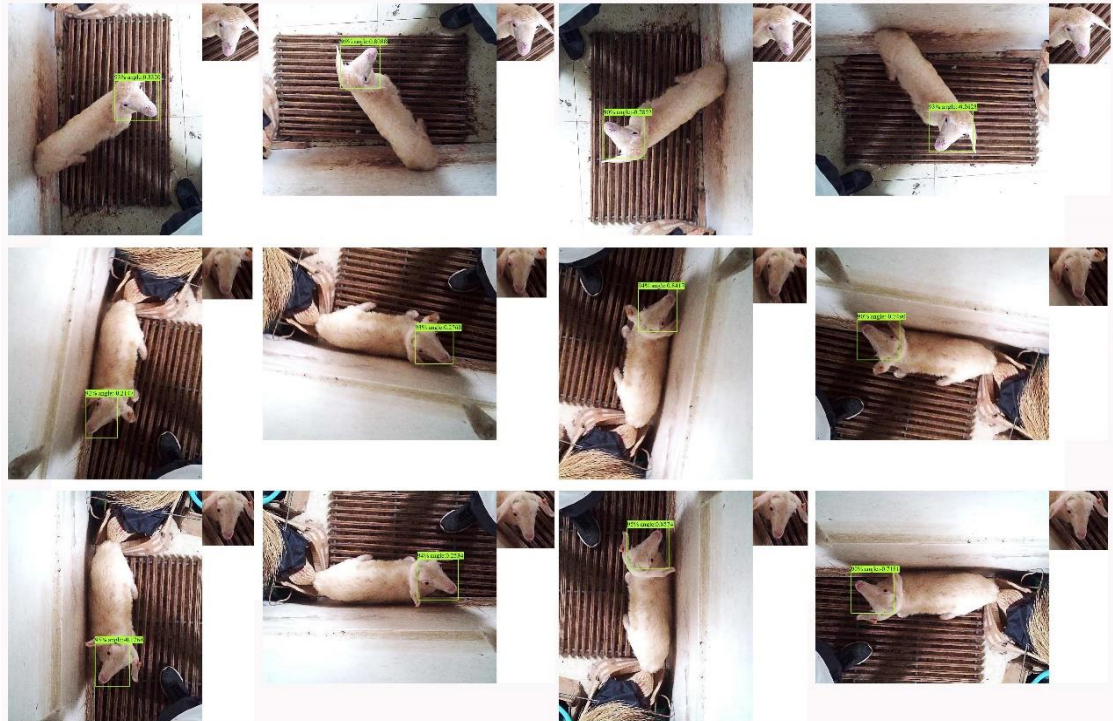

Fig. 6. Detection examples on sheep bird-view image with LAD-RCNN. The small image in the upper right corner of each image is the extracted normalized face according to the detection result. The four pictures in each line represent the same picture in test set, which are original image and images rotated by 90°, 180 ° and 270°, respectively

## 4.4 Experiments on goat infrared image dataset

The self-made goat infrared image dataset labeled the location of the goat face and the rotation angle of goat face, which contains 2409 training data and 1000 test data. The datasets 1 containing angle information and datasets 2 without angle information were both generated from all the training data.

The training parameters were set as follow: the probability of horizontal flip and 90° rotation of dataset 1 were set to 0.5, and the probability of vertical rotation of dataset 1 was set to 0.55; The probability of horizontal flip of dataset 2 was set to 0.5; The probability of 90° rotation and vertical rotation of dataset 2 were set to 0; The probability of 2×2 merger in dataset 1 and dataset 2 were set to 0.8; The batchsize of dataset 1 was set to 7, and that of dataset 2 was set to 5; The input image channel was set to 1; The total training step was set to 50000.

Table 3. Test result of LAD-RCNN on goat infrared image dataset

| Backbone | Precision | Recall | F1 score | AP | AAD |
| --- | --- | --- | --- | --- | --- |
| Ours | 96.43% | 98.39% | 97.40% | 98.19% | 4.62° |
| MobileNetV2 | 97.20% | 97.66% | 97.43% | 98.35% | 4.96° |
| VGG16 | 89.95% | 96.69% | 93.20% | 96.30% | 5.94° |
| ResNet50 | 96.93% | 98.83% | 97.87% | 98.29% | 4.48° |

Note: AAD represents average angle difference between the detection angle and the ground-truth angle.

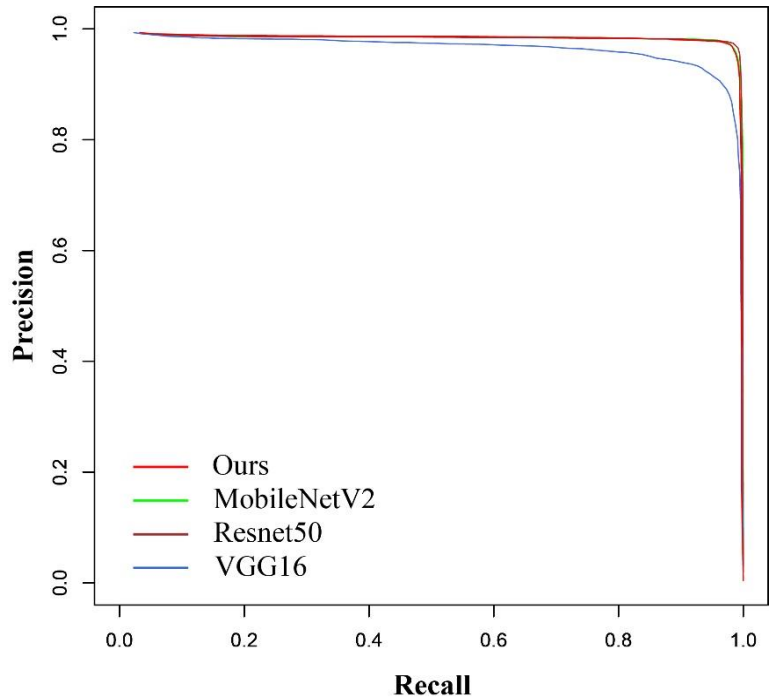

Fig. 7. Precision-Recall Curves on goat infrared image dataset

To evaluate the performance of LAD-RCNN on detecting goat face with arbitrary direction in infrared image, we rotated the test image by 90°, 180° and 270° respectively to form a new test dataset with 4000 images. The test results (Table 3, Fig. 7 and Fig.8.) show that all the AP were more than 96%, and all the average angle difference were within 5.94°. When Ours, MobileNetV2 or ResNet50 were adopted as the backbone network, the AP were more than 98% and the average angle difference were within 4.96°.

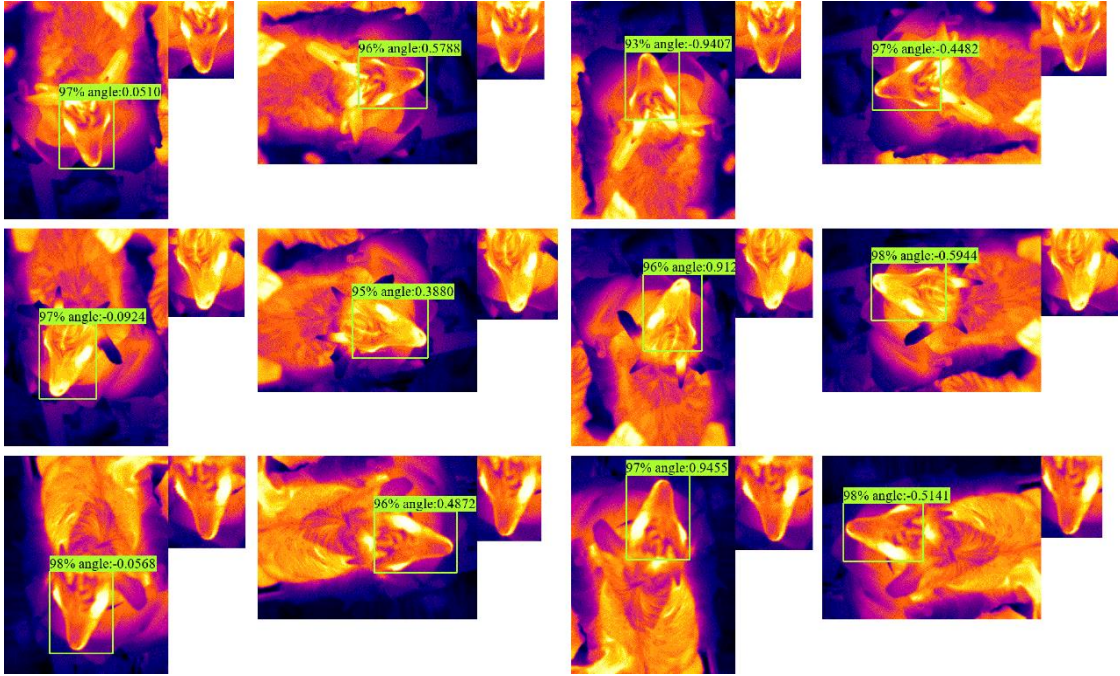

Fig.8. Detection examples on goat infrared image with LAD-RCNN. The small image in the upper right corner of each image is the extracted normalized face according to the detection result. The four pictures in each line represent the same picture in test set, which are original image and images rotated by 90°, 180 ° and 270°, respectively

## 4.5 Experiments on other dataset

To test the performance of LAD-RCNN in various datasets, we evaluated LAD-RCNN on Celeba dataset and lfw_5990 datasets respectively.

The Celeba dataset contains 202599 pictures. The location of the face and two eyes are labeled in Celeba dataset [38]. The rotation angle of face was calculated according to the two eyes location. We randomly selected 192599 pictures in Celeba dataset as training data and 10000 pictures as test data. The dataset 1 containing angle information was generated by training data in Celeba dataset; The dataset 2 without angle information was generated by WIDER dataset.

The training parameters were set as follows: the probability of horizontal flip, vertical flip and 90° rotation of dataset 1 were set to 0.5; The probability of 2×2 merger and 3×3 merger of dataset 1 were set to 0.3. All the all the transformation parameters of dataset 2 were set to 0. The batchsize of dataset 1 was set to 7, and that of dataset 2 was set to 5; The input image channel was set to 3; The total training step was set to 25000.

The test results are shown in Table 4. All the AP were more than 98%, and all the average angle difference were within 5.19°.

Table 4.  Test result of LAD-RCNN on Celeba dataset

| Backbone | Precision | Recall | F1score | AP | AAD |
|---|---|---|---|---|---|
| Ours | 97.75% | 96.82% | 97.28% | 98.62% | 5.05° |
| MobileNetV2 | 97.85% | 96.97% | 97.41% | 98.57% | 5.19° |
| VGG16 | 97.54% | 97.38% | 97.46% | 98.46% | 4.83° |
| ResNet50 | 98.46% | 97.11% | 97.78% | 98.76% | 4.79° |

Note: AAD represents average angle difference between the detection angle and the ground-truth angle.

The lfw dataset contains 5590 pictures, in which 4149 pictures are in training set and 1437 pictures are in test set. The location of the face and two eyes are labeled in lfw_5590[39]. The rotation angle of face was calculated according to the two eyes location. The dataset 1 containing angle information was generated by training data in lfw_5590 dataset; The dataset 2 without angle information was generated by WIDER dataset.

The training parameters were same as the training parameters of Celeba dataset. The test results are shown in Table 5. All the AP were more than 94%, and all the average angle difference were within 4.36°. When Ours, MobileNetV2 or ResNet50 were adopted as the backbone network, the AP were more than 95% and the average angle difference were within 3.53°.

Table 5. Test result of LAD-RCNN on lfw_5590 dataset

| Backbone | Precision | Recall | F1score | AP | AAD |
| --- | --- | --- | --- | --- | --- |
| Ours | 99.74% | 90.96% | 95.15% | 95.32% | 2.99° |
| MobileNetV2 | 99.76% | 91.35% | 95.37% | 95.13% | 3.53° |
| VGG16 | 99.41% | 90.13% | 94.54% | 94.05% | 4.36° |
| ResNet50 | 99.67% | 92.54% | 95.97% | 96.33% | 2.68° |

Note: AAD represents average angle difference between the detection angle and the ground-truth angle.

## 5 Discussion

Compared with the two-stage method, the one-stage object detector gets rid of the time-consuming regional proposal step and directly detect object from the densely predesigned candidate boxes, which has faster detecton speed [27]. Lin et al [16] propose focal loss to soloves the problem of "imbalance between positive and negative samples" in one-stage object detector, and so that the one-stage detector achieve well performance in rotation object detection [40]. In addition, in filed of livestock face recognition, it is no need to detect the too small objects. That is, it is only need to detect the face and its direction with normal size in filed of livestock face recognition. Therefore, LAD-RCNN was designed by one-stage strategy.

LAD-RCNN consists of backbone, neck network and head network (Fig. 3). LAD-RCNN supports the use of various backbone network. We also designed a lightweight backbone for LAD-RCNN. The evaluation results on multiple datasets show that when using LAD-RCNN with our backbone to detect face with arbitrary directions, the AP was more than 95%, the average angle difference between the detection angle and the ground-truth angle were within 6.42° (Table 2-Table 5). The backbone evaluation result show that the number of parameters in our backbone is 5.21 times and 8.36 times less than that in VGG16 and ResNet50 respectively, and the detection speed of our backbone is 47%, 104% and 150% faster than MobileNetV2, VGG16 and ResNet50 respectively. Therefore, the backbone proposed in this study improves the detection speed without reducing the detection accuracy and is very suitable for LAD-RCNN.

Infrared thermal imaging technology is a fast non-contact temperature measurement technology, which can generate images based on the surface temperature information and provide dynamic information of surface temperature changes caused by physiological processes. It has been widely used in animal research [41-45]. Based on the characteristics of infrared images, we speculate that animal recognition in infrared images will become one of the research hotspots. In order to adapt LAD-RCNN to infrared images with single channel, we add a channel number configuration interface in config file of LAD-RCNN. LAD-RCNN will adapt to infrared thermal image if the channel number is set to 1. The test results on goat infrared image (Table 3, Fig. 7, Fig.8) shows that LAD-RCNN performs well in face detection on infrared images.

In the field of animal research, small dataset may be required to be used for face identification for some reasons [46, 47]. In order to perform better in small datasets, LAD-RCNN integrates some dataset enhancement functions, such as horizontal flip, vertical flip ,90° rotation, 2×2 merger and 3×3 merger. The training set of goat dataset only contains 1680 data, of which only 438 data contain rotation angle information. The evaluation result on this dataset show that the AP reached 97.55% and the average angle difference between the detection angle and the ground-truth angle was within 6.42°, which proves that LAD-RCNN performs well in the small dataset.

It is a pity that no more livestock datasets have been found for extensive verification of LAD-RCNN due to most of the livestock recogniztion studies have not published their labeled dataset. To test the extensive applicability of LAD-RCNN in different datasets, we further tested LAD-RCNN by human dataset Celeba and lfw_5990. The evaluation result in multiple dataset proves the extensive applicability of LAD-RCNN in various datasets. The experimental conditions tested on all dataset has been reported in detail on this paper. Peers of livestock face detection research can directly employ LAD-RCNN in their study to realize face detection and normalization.

# 6 Conclusions

We propose a Light-weight Angle Detection and Region-based Convolutional Network (LAD-RCNN) for livestock face detection and normalization, which can detect the livestock face and rotation angle with arbitrary directions in one-stage. The backbone proposed by this study is a light-weight network, and the detection speed of our backbone is 72.74FPS, which is faster than that of MobileNetV2, VGG16 and ResNet50. LAD-RCNN has been evaluated on multiple datasets and the AP was more than 95%, the average angle difference between the detection angle and the ground-truth angle was within 6.42°. LAD-RCNN performs well on small dataset and infrared image with single channel. This shows that the LAD-RCNN has excellent performance in livestock face detection and angle based normalization.

# Supplementary Figures

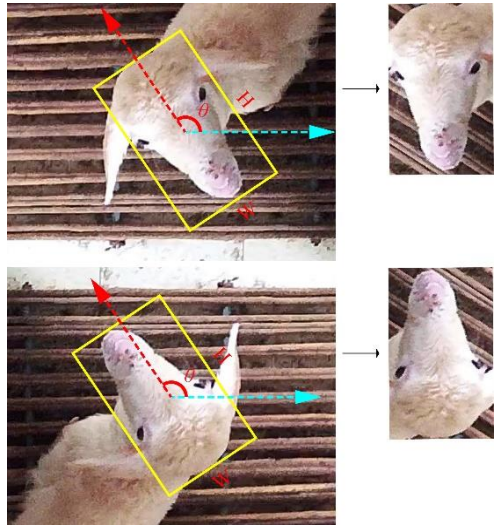

Supplementary Fig. 1 Angle encoding method in other study. The angle is represented as the angle between the long axis and the horizontal axis. In this way, an inverted face image may be obtained. Therefore, the angle encoding method in other study is not suitable for animal face recognition and normalizing.